\documentclass[10pt,twocolumn,letterpaper]{article}

\usepackage[pagenumbers]{cvpr} 

\usepackage{arydshln}
\usepackage{array}
\usepackage{booktabs}
\usepackage{multirow}
\usepackage{makecell}
\newcommand{\tabincell}[2]{\begin{tabular}{@{}#1@{}}#2\end{tabular}}

\definecolor{cvprblue}{rgb}{0.21,0.49,0.74}
\usepackage[pagebackref,breaklinks,colorlinks,allcolors=cvprblue]{hyperref}

\def\paperID{7120} 
\def\confName{CVPR}
\def\confYear{2025}

\title{3D Scene Graph Guided Vision-Language Pre-training}

\author{Hao Liu\textsuperscript{1} \quad Yanni Ma\textsuperscript{2} \quad Yan Liu\textsuperscript{2} \quad Haihong Xiao\textsuperscript{3} \quad Ying He\textsuperscript{1}\\
\textsuperscript{1}Nanyang Technological University \quad \textsuperscript{2}Sun Yat-Sen University \\ \textsuperscript{3}South China University of Technology\\
}

\begin{document}
\maketitle
\begin{abstract}

3D vision-language (VL) reasoning has gained significant attention due to its potential to bridge the 3D physical world with natural language descriptions. Existing approaches typically follow task-specific, highly specialized paradigms. Therefore, these methods focus on a limited range of reasoning sub-tasks and rely heavily on the hand-crafted modules and auxiliary losses. This highlights the need for a simpler, unified and general-purpose model. In this paper, we leverage the inherent connection between 3D scene graphs and natural language, proposing a 3D scene graph-guided vision-language pre-training (VLP) framework. Our approach utilizes modality encoders, graph convolutional layers and cross-attention layers to learn universal representations that adapt to a variety of 3D VL reasoning tasks, thereby eliminating the need for task-specific designs. The pre-training objectives include: 1) Scene graph-guided contrastive learning, which leverages the strong correlation between 3D scene graphs and natural language to align 3D objects with textual features at various fine-grained levels; and 2) Masked modality learning, which uses cross-modality information to reconstruct masked words and 3D objects. Instead of directly reconstructing the 3D point clouds of masked objects, we use position clues to predict their semantic categories. Extensive experiments demonstrate that our pre-training model, when fine-tuned on several downstream tasks, achieves performance comparable to or better than existing methods in tasks such as 3D visual grounding, 3D dense captioning, and 3D question answering.

\end{abstract}    
\section{Introduction}
\label{sec:intro}

3D vision-language (VL) reasoning is an emerging research field that seeks to connect the 3D physical world with natural language descriptions, with broad applications such as human-machine interaction and embodied intelligence. It requires algorithms to align visual information with textual descriptions, enabling a shared feature space for both modalities. Recently, 3D VL reasoning has received increasing attention, with several methods proposed to tackle various tasks, including 3D visual grounding (VG) \cite{ScanRefer,Referit3d}, 3D dense captioning \cite{Scan2Cap} and 3D question answering \cite{ScanQA}.

Despite significant advancements in handling these 3D VL tasks, most of existing methods remain highly specialized and tailored to specific tasks. Typically, they focus on one or two tasks and rely heavily on the design of complex modules and auxiliary losses. For instance, 3D-SPS \cite{3D-SPS} introduces a language-aware down-sampling module to select target-related keypoints, while MVT \cite{MVTransformer} and ViewRefer \cite{ViewRefer} use point cloud rotation and multi-view positional encoding to create a view-robust representation. In contrast, current mainstream 2D approaches are based on the vision-language pre-training (VLP) framework. They are generally pre-trained on large-scale image-text pairs, then fine-tuned to adapt to various downstream tasks. The pre-training process facilitates learning universal representations with strong transferability through contrastive learning \cite{SimCLR,CLIP} and masked modality learning \cite{MaskedMAE}. However, 3D VLP is still in its infancy due to the unique properties of 3D point clouds and the misalignment between 3D point clouds and natural language descriptions.

\begin{figure}[t]
    \centering
    \includegraphics[width=\columnwidth]{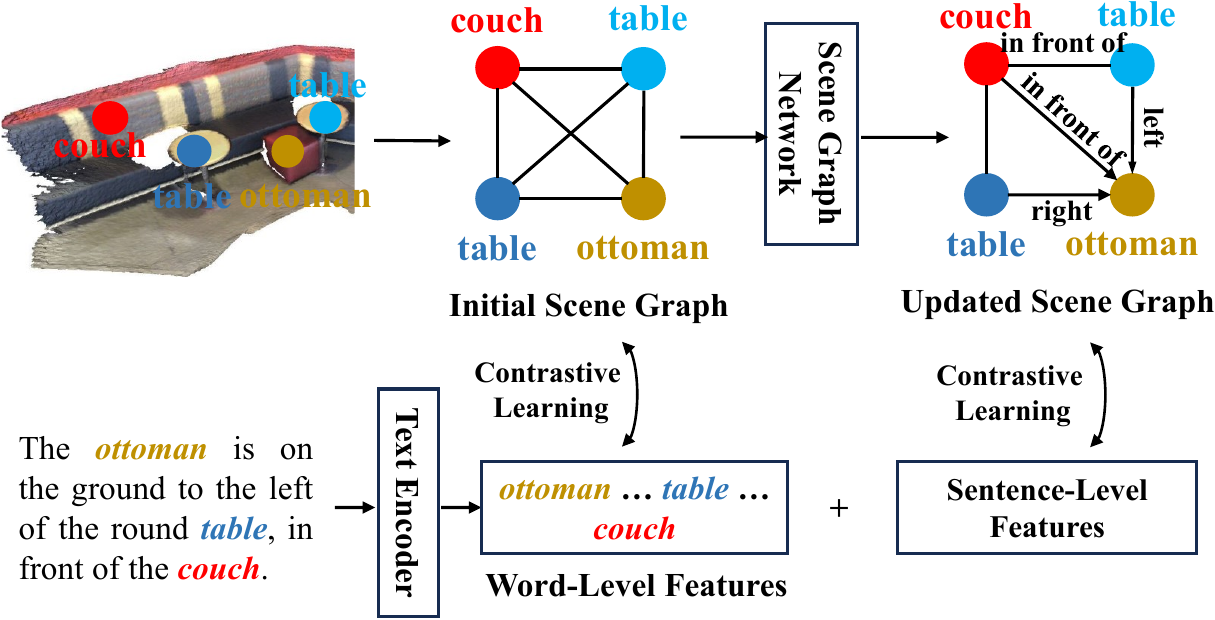}
    \caption{An illustration of the natural alignment between 3D scene graphs and natural language descriptions. We leverage this correspondence to pre-train our 3D-language model in the form of contrastive learning.}
    \label{fig_1}
\end{figure}
On the other hand, a handful of studies \cite{FFL-3DOG} have begun to explore 3D scene graph \cite{3DSSG,ma2025heterogeneous}, which model both objects and their relationships, to address the 3D VG task. We observe that 3D scene graphs naturally align with natural language descriptions. As shown in Fig. \ref{fig_1}, a subject (such as ``ottoman''), a predicate (such as ``left'') and an object (such as ``table'') form the basic components of any sentence, whereas in a scene graph, the same $<$subject, predicate, object$>$ triplet can be represented by two nodes and one edge. As a result, an object name (word-level) may correspond to multiple objects (nodes) of the same category in the initial stage of the scene graph, while a sentence ultimately corresponds to a specific node (\textit{i.e.}, the referential target) after scene graph learning. This natural correspondence makes cross-modality contrastive learning highly effective. Therefore, a natural question arises: \emph{Can we leverage the relationship between 3D scene graphs and natural language descriptions to design a 3D VLP scheme?} We answer this in the affirmative and demonstrate that our scene graph-guided pre-training model outperforms task-specific methods across various 3D VL reasoning tasks.

In this paper, we propose a 3D scene graph-guided vision-language pre-training scheme that establishes multi-level alignments between 3D objects and input text to learn universal representations for various 3D VL reasoning tasks. First, inspired by the close relationship between scene graphs and natural language descriptions, we propose a scene graph-guided multi-level contrastive learning (SG\_MCL) strategy. Different from the global contrastive learning in CLIP \cite{CLIP}, our method aligns 3D objects with textual features at multiple fine-grained levels, \textit{e.g.}, \emph{word-object level}, \emph{sentence-referred object level} and \emph{scene-level}. Next, we introduce a masked modality learning (MML) approach that reconstructs the masked portions of the input modality to improve the generalization of feature representation. Due to the sparsity and irregularity of 3D point clouds, we predict the semantic categories of masked objects by leveraging positional clues along with the remaining 3D objects and text, rather than directly reconstructing their 3D points. The pre-training model is built primarily on simple, general-purpose modules to learn transferable multi-modal features, thus eliminating the need for complex, task-specific modules and losses. The main contributions of this work are summarized as follows:
\begin{itemize}
    \item We propose a novel 3D vision-language pre-training framework that uses simple, universal modules to learn transferable multi-modal features without any task-specific designs.
    \item We propose a scene graph-guided contrastive learning strategy that leverages the strong alignment between 3D scene graphs and language to match 3D objects with textual features across different fine-grained levels.
    \item Extensive experiments on ScanRefer \cite{ScanRefer}, Scan2Cap \cite{Scan2Cap} and ScanQA \cite{ScanQA} demonstrate that our pre-training model achieves competitive performance across multiple 3D VL tasks after fine-tuning.
\end{itemize}
\section{Related works}
\label{sec:related_works}

\begin{figure*}
    \centering
    \includegraphics[width=\linewidth]{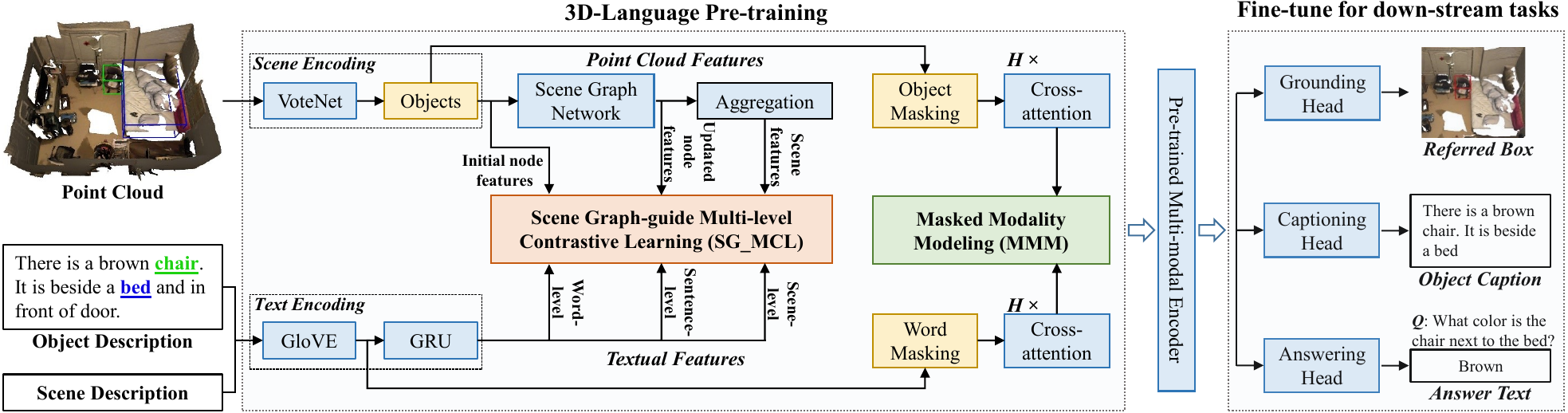}
    \caption{The overview of our model. Given a 3D point cloud-text pair, we first use a scene encoding module to extract 3D object proposals and a text encoder to generate textual features. We then treat the 3D object proposals as nodes to construct the scene graph, using a scene graph learning module to update node and edge features. Finally, the model is pre-trained with the proposed scene graph-guided multi-level contrastive learning and masked modality modeling. The pre-trained model can be fine-tuned for various downstream tasks, including 3D visual grounding, 3D dense captioning and 3D question answering.}
    \label{fig_2}
\end{figure*}

\subsection{Vision-language pre-training}

Vision-Language Pre-training (VLP) \cite{CLIP,ALIGN,VL-Bert,PointCLIP,PointCLIPv2,CLIP2Point} aims to leverage contrastive learning to learn universal representations that enhance performance on down-stream tasks. Large-scale image (or point cloud)-text pairs are essential to the success of VLP. Recently, CLIP \cite{CLIP} and ALIGN \cite{ALIGN} have received significant attention due to their superior feature transferability and cross-modal understanding capabilities. These models perform contrastive learning on massive web-crawled image-text pairs to create a unified embedding space for both image and text features. Yang et al. \cite{yang2022vision} introduced intra-modal supervision to enhance feature learning within each modality, while Duan et al. \cite{duan2022multi} proposed an innovative codebook to better align multi-modal representations.

Despite remarkable progress in 2D VLP, extending these techniques to the 3D domain is highly challenging due to the unique characteristics of point clouds and the scarcity of available point cloud-text pairs. To address this, Zhang et al. \cite{PointCLIP} projected point clouds into multi-view depth images and then used pre-trained CLIP for zero-shot or few-shot 3D classification. Zhu et al. \cite{PointCLIPv2} introduced a realistic projection to generate CLIP-preferred depth images and prompted GPT \cite{GPT} to create 3D-specific texts. Chen et al. \cite{UniT3D} leveraged the natural correspondences between 2D and 3D by feeding 2D image crops into an image captioner to generate descriptions and extracting 3D frustums of these crops, thus producing rich point cloud-text pairs. Jin et al. \cite{3D-VLP} pre-trained their model using word-region (object) alignment and a masked modeling strategy.

\subsection{3D vision and language reasoning}

\textbf{3D visual grounding (VG)} aims to locate a target object within point clouds according to a free-form language description. Chen et al. \cite{ScanRefer} introduced the first dataset, \emph{ScanRefer}, and proposed a two-stage VG framework. This framework initially employs a 3D detector to obtain object proposals and then selects the best-matching proposal by integrating language features. Yuan et al. \cite{InstanceRefer} used the predicted target category to filter out redundant proposals, while Zhao et al. \cite{3DVG-Transformer} applied transformer-like modules to model proposal relations, helping to distinguish the target object from similar ones. In contrast, Luo et al. \cite{3D-SPS} proposed a single-stage method, \emph{3D-SPS}, which frames 3D VG as a keypoint selection problem. Specifically, it identifies a set of keypoints guided by the input text and uses cross-modal attention layers to locate the target object by aligning keypoint features with language features. Wu et al. \cite{EDA} parsed semantic components from the input text and then aligned visual features with the parsed component features to achieve fine-grained visual-text fusion.

\textbf{3D dense captioning} involves locating and describing objects of interest within a 3D scene. Chen et al. \cite{Scan2Cap} proposed the pioneering work \emph{Scan2Cap}, which first detects objects in the scene and then describes detected objects using an attention-based captioning module. Jiao et al. \cite{MORE} generated comprehensive captions by fully mining complex object relationships. Similarly, Wang et al. \cite{SpaCap3D} introduced relative spatiality modeling into a Transformer network, \textit{i.e.,} using spatial relationships to enhance vision tokens. Zhong et al. \cite{Contextual} incorporated superpoints into the network to supplement contextual information, such as non-object details. Chen et al. \cite{Vote2Cap-DETR} encoded the input scene into a set of vote queries, which are then processed by a Transformer decoder with parallel detection and captioning heads.

\textbf{3D question answering (QA)} requires algorithms to answer a given question and locate question-relevant objects. ScanQA \cite{ScanQA} first encodes the question into language features and extracts object proposals from the 3D scene. It then feeds the combined proposal and question features into an object localization module and an answer classification module. Parelli et al. \cite{Clip-Guided} and Delitzas et al. \cite{Multi-CLIP} aligned 3D extracted features with corresponding captions and 2D images in the CLIP embedding space, thus incorporating CLIP’s prior knowledge into the network. In this work, we evaluate our pre-trained model on these 3D VL tasks.

\subsection{3D scene graph}
3D scene graph (SG) is a compact 3D scene representation that models both objects (nodes) and their relationships (edges) within a scene. Wald et al. \cite{3DSSG} utilized the PointNet \cite{PointNet} network to extract object features and then applied graph convolutional networks (GCNs) to predict node and edge categories in the scene graph. Lv et al. \cite{SGFormer} proposed the Semantic Graph Transformer (SGFormer) framework, which incorporates prior knowledge from large language models (LLMs) to enhance object features. Wang et al. \cite{VL-SAT} leveraged visual-linguistic information from 2D images and the CLIP \cite{CLIP} model to assist in training the 3D scene graph network. Koch et al. \cite{Lang3DSG} proposed a language-based contrastive learning strategy to distill CLIP \cite{CLIP} knowledge into the network. In this work, we leverage the inherent alignment between scene graphs and language descriptions to establish multi-level alignment for 3D VL pre-training.
\section{Methodology}
\label{sec:method}

\subsection{Overview}

We leverage the natural alignment between language descriptions and 3D scene graphs to design a 3D visual-language pre-training (VLP) scheme based on multi-level contrastive learning. As shown in Fig. \ref{fig_2}, our pre-training model consists of four main modules: a scene encoding module, a text encoding module, a scene graph learning module, and a cross-modality fusion module.

\textbf{Scene encoding.} The scene encoder takes point clouds $P_{input}\in{\mathbb{R}^{N\times{(3+F)}}}$ with 3D \emph{xyz} coordinates and $F$-dimension features as input, where $N$ is the number of point clouds. We use VoteNet \cite{VoteNet} with a PointNet++ \cite{PointNet++} backbone to extract $K$ seed points $P_{seed}\in{\mathbb{R}^{K\times{(3+C_p)}}}$ and generate $M$ 3D object proposals $F_o\in{\mathbb{R}^{M\times{C_p}}}$, where $C_p$ is the feature dimension.

\textbf{Text encoding.} The input text $T_{input}$ is first encoded into 300-dimensional embedding vectors $W=\{w_i\}_{i=1}^L\in{\mathbb{R}^{L\times{300}}}$ using pre-trained GloVE \cite{GloVE}, where $L$ is the length of the input text. We then feed the embedding vectors $W$ into a GRU \cite{GRU} cell to obtain word-level features $F_w\in{\mathbb{R}^{L\times{C_l}}}$ and sentence-level features $F_s\in{\mathbb{R}^{C_l}}$.

\textbf{Scene graph learning.} We first create a 3D scene graph $\mathcal{G=(V,E)}$, where the nodes $\mathcal{V}$ represent object proposals and the edges $\mathcal{E}$ denote relationships between proposals. The scene graph $\mathcal{G}$ is then processed by the scene graph network to update the node features. Finally, graph pooling is applied to aggregate graph nodes $\mathcal{V}$ to obtain a scene-level representation. We leverage the natural alignment between the input text and 3D scene graph to perform multi-level contrastive learning (See Section \ref{MMCL}).

\textbf{Cross-modality fusion.} We introduce masked modality modeling for 3D vision-language pre-training. The remaining word features and object features are fused through $H$ cross-modality attention layers to reconstruct the missing words and object proposals (See Section \ref{MMM}).

\subsection{Scene graph-guided multi-level contrastive learning\label{MMCL}}

\begin{figure}[t]
    \centering
    \includegraphics[width=\columnwidth]{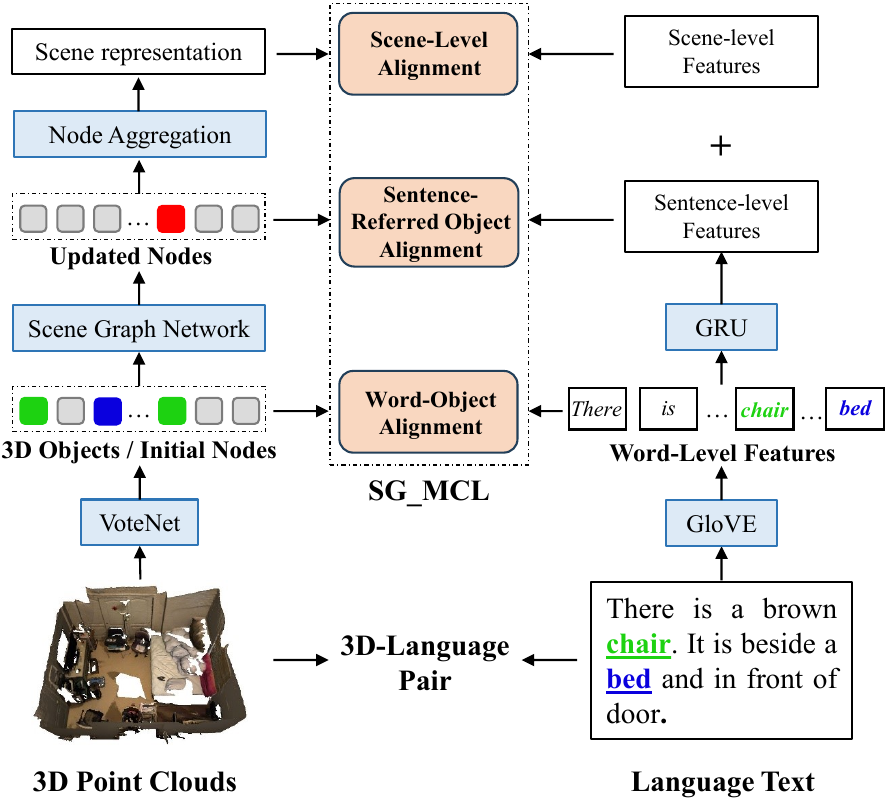}
    \caption{Scene graph-guided multi-level contrastive learning (SG\_MCL) strategy. It aligns 3D object and textual features at various levels, i.e., word-object level, sentence-referred object level and scene-level.}
    \label{fig_3}
\end{figure}

3D scene graph is an emerging scene representation that models objects in a 3D scene as well as their relationships. It is observed that 3D scene graphs naturally align with language descriptions. In language, a subject, a predicate and an object form the fundamental components of a sentence, while in a scene graph, the same $<$subject, predicate, object$>$ triplet is represented by two nodes and an edge. Thus, an object name (a word) may correspond to multiple objects (nodes) of the same category in the initial stage of scene graph, while a sentence corresponds to a specific node (i.e., the referential target). Leveraging this natural correspondence, we design a multi-level alignment strategy based on scene graphs, aligning 3D objects and language features at different fine-grained levels: \emph{word-object level}, \emph{sentence-referred object level} and \emph{scene-level}, as shown in Fig. \ref{fig_3}.

\textbf{Level 1 - word-object alignment.} We construct an initial scene graph $\mathcal{G=(V,E)}$, where $\mathcal{V}$ represents the 3D object proposals, and $\mathcal{E}$ denotes their relationships. Before feeding $\mathcal{G}$ into the scene graph network, we perform fine-grained alignment between the 3D object proposals $O$ and word-level features $F_w$, \textit{i.e.}, aligning each 3D object proposal with its corresponding object name (\textit{i.e.}, semantic categories). For example, given the sentence \emph{``there is a black \underline{chair} next to the \underline{cabinet}"}, we first parse the object names (such as \emph{\underline{chair}} and \emph{\underline{cabinet}}) and then find all object proposals associated with each parsed object name. Notably, we focus only on the object names of the referential object and the auxiliary object. Finally, a binary cross-entropy loss is used to align the 3D object proposals with the word features located at the positions of the corresponding object names:
\begin{align}
    L_{WO} = & -\frac{1}{B \cdot N \cdot L_{b}} \sum_{i=1}^B \sum_{j=1}^N \sum_{k=1}^{L_b} \Big[ 
    \, \mathbf{s}_{jk}^{i} \cdot \mathrm{log} \, \sigma(\mathbf{o}^i_{jk} \cdot \mathbf{w}^i_{jk}) \nonumber \\
    & + (1 - \mathbf{s}_{jk}^{i}) \cdot \mathrm{log}(1 - \sigma(\mathbf{o}^i_{jk} \cdot \mathbf{w}^i_{jk})) \Big]
\end{align}
where $\sigma$ is the sigmoid function, $B$ is the training batch size, and $L_b$ is the length of input text in batch $b$. $\mathbf{o}^i_{jk}\cdot{\mathbf{w}^i_{jk}}$ denotes the predicted similarity score between the $j$-th object proposal and the $k$-th word in batch $i$, while $\mathbf{s}_{jk}^{i}$ represents the ground truth similarity score, ranging from 0 to 1. $\tau$ is the temperature parameter. At the initial stage of the scene graph, one object name may correspond to multiple 3D object proposals.

\textbf{Level 2 - sentence-referred object alignment.} We treat the 3D object proposal features $F_o$ as initial node features $\phi_{n}^0$. For initial edge features $\phi_{e}^0$, we only consider the $N_1$ neighboring object proposals for each proposal, focusing on the relationships between the referential object and its neighboring (or auxiliary) objects. The main object component, spatial relationship component and auxiliary object component in the sentence are arranged as a triplet $t_{ij}=<\phi_{n,i}^0,\phi_{e_{ij}}^0,\phi_{n,j}^0>$. To align sentence-level textual features, we employ a scene graph network (\textit{i.e.}, several graph convolutional layers with message passing) to propagate information through the graph, allowing each node and edge to incorporate contextual information from its neighbors. Taking the $l_m$-th graph convolution layer as an example, we illustrate the process of updating node and edge features. First, we feed the triplet $t_{ij}^{(l_m)}$ into a multi-layer perceptron (MLP) $g_1(\cdot)$:
\begin{equation}
(\psi_{n,i}^{(l_m)},\phi_{e_{ij}}^{(l_m+1)},\psi_{n,j}^{(l_m)})=g_1(\phi_{n,i}^{(l_m)},\phi_{e_{ij}}^{(l_m)},\phi_{n,j}^{(l_m)})
\end{equation}
where $\psi_{n,i}^{(l_m)}$ are the incoming features of the $i$-th node, and $\phi_{e_{ij}}^{(l_m+1)}$ are the updated edge features. The updated node features $\phi_{n,i}^{(l_m+1)}$ are calculated as follows:
\begin{equation}
    \phi_{n,i}^{(l_m+1)}=\phi_{n,i}^{(l_m)}+g_2\left(\frac{1}{N_i}\left(\sum_{k\in{\mathcal{R}_i}}\psi_{n,k}^{(l_m)}+\sum_{k\in{\mathcal{R}_j}}\psi_{n,k}^{(l_m)}\right)\right)
\end{equation}
where $N_i$ is the number of nodes connected to node $i$, and $\mathcal{R}_i$ and $\mathcal{R}_j$ are the sets of nodes connected to node $i$ and node $j$, respectively. After processing through the scene graph network, the updated node and edge features contain contextual information about their neighbors. Consequently, we align the updated node features $\phi_n$ with the sentence-level features $F_s$ to perform sentence-referred object alignment as follows: 
\begin{equation}
    L_{SRO}=-\frac{1}{B}\sum_{i=1}^B\left[\mathrm{log}\frac{\mathrm{exp}
    (\hat{\phi}_n^i\cdot{F_s^i}/\tau)}{\sum_{j=1}^B\mathrm{exp}(\hat{\phi}_n^i\cdot{F_s^j}/\tau)}\right]
\end{equation}
where $F_s^i$ represents sentence-level features, and $\hat{\phi}_n^i$ denotes the node features corresponding to the referential object.

\textbf{Level 3 - scene-level alignment.} After obtaining the updated node features, we aggregate all node features to generate a scene-level representation $F_{sc}$, \textit{i.e.}, $F_{sc}=\mathrm{Agg}(\phi_{n})$. We then align the scene-level features $F_{sc}$ and the textual features $F_{des}$ of the scene description to achieve scene-level alignment:
\begin{equation}
    L_{Scene}=-\frac{1}{B}\sum_{i=1}^B[\mathrm{log}\frac{\mathrm{exp}(F_{sc}^i\cdot{F_{des}^i/\tau})}{\sum_{j=1}^B\mathrm{exp}(F_{sc}^i\cdot{F_{des}^j/\tau})}]
\end{equation}
The loss for scene graph-guided multi-level contrastive learning is given by $L_{SG\_MCL}=L_{WO}+L_{SRO}+L_{Scene}$.

\subsection{Masked modality modeling \label{MMM}}

After aligning 3D object features with textual features at various fine-grained levels, we can fine-tune the model for several 3D vision-language reasoning tasks. To achieve a comprehensive understanding and meaningful bidirectional interaction between modalities, we introduce masked modality modeling to jointly pre-train the model. 

\textbf{Masked language modeling (MLM).} Following the pre-training methods of large language models like BERT \cite{BERT}, we perform MLM by randomly selecting a subset of input words and replacing them with the ``unk'' token. The masked word features $F_w^{mask}$ and 3D object features $F_o$ are fed into a cross-attention module, followed by three MLP layers to predict the masked words:
\begin{equation}
    \hat{y}_w=\mathrm{MLP}(\mathrm{CrossAttn}(F_w^{mask},F_o,F_o))
\end{equation}
where $\hat{y}_w$ represents the predicted masked words (i.e., vocabulary probabilities). The model is trained using cross-entropy loss:
\begin{equation}
    L_{MLM}=\mathrm{CrossEntropy}(\hat{y}_w,y_w)
\end{equation}
where $y_w$ are the ground truth words for the masked tokens.

\textbf{Masked object modeling (MOM).} Similar to MLM, we randomly mask a portion of the 3D object proposals and replace their features with mask tokens $M$ (\textit{i.e.}, a set of learnable parameters). The visible object features $F_o^{vis}$ and mask tokens $M$ are concatenated and augmented with positional embeddings $e_o$ to form full token set $S_o=\mathrm{Concat}(F_o^{vis},M)+e_o$. The positional embeddings $e_o$ are generated by applying a linear layer to the 27-dimensional object position attributes (including the box center and eight box corners). We feed the full token set $S_o$ and word-level features $F_w$ into a cross-attention module, followed by three MLP layers, to predict the semantic categories $\hat{c}_o$ of the 3D object proposals:
\begin{equation}
    \hat{c}_o=\mathrm{MLP}(\mathrm{CrossAttn}(S_o,F_w,F_w))
\end{equation}
The MOM loss is calculated as follows:
\begin{equation}
    L_{MOM}=\mathrm{CrossEntropy}(\hat{c}_o^{m},c_o^{m})
\end{equation}
where $c_o$ are the ground-truth semantic labels. Note that supervision is applied only to the masked tokens. The total loss for masked modality modeling, $L_{MMM}$, is given by $L_{MMM}=L_{MLM}+L_{MOM}$.

\subsection{Pre-training objectives}

In addition to the aforementioned $L_{SG\_MCL}$ and $L_{MMM}$, we also incorporate a detection loss $L_{DET}$ \cite{VoteNet} and a language-to-object classification loss $L_{lang}$ \cite{ScanRefer} to jointly train the model. Thus, the overall pre-training loss is a weighted sum of all these losses: $L_{pre}=a\cdot{L_{SG\_MCL}}+b\cdot{L_{MMM}}+c\cdot{L_{DET}}+d\cdot{L_{lang}}$.

\textbf{Detection loss.} We use the detection loss proposed in VoteNet \cite{VoteNet} to jointly optimize the detection module. The detection loss $L_{DET}$ is composed of the vote regression loss $L_{vote}$, the objectness classification loss $L_{obj}$, the box regression loss $L_{box}$, and the semantic classification loss $L_{sem}$.

\textbf{Language-to-object classification loss.} Following previous methods \cite{ScanRefer}, we introduce a language-to-object classification loss $L_{lang}$ to supervise language-based object classification.
\section{Experiments}
\label{sec:experiments}

\begin{table*}[t]
    \centering
    \resizebox{\textwidth}{!}{
    \begin{tabular}{c|r|c|c|c|cc|cc|cc}
        \Xhline{2.0\arrayrulewidth}
        & \multirow{2}{*}{Method} & \multirow{2}{*}{Detector} & \multirow{2}{*}{Dataset} & \multirow{2}{*}{Input} & \multicolumn{2}{c|}{Unique} & \multicolumn{2}{c|}{Multiple} & \multicolumn{2}{c}{Overall}  \\
         & & & & & Acc@0.25 & Acc@0.5 & Acc@0.25 & Acc@0.5 & Acc@0.25 & Acc@0.5 \\
        \Xhline{2.0\arrayrulewidth}
        \multirow{14}{*}{\rotatebox{90}{Task-specific}} & TGNN \cite{TGNN} & PointGroup & ScanRefer & 3D & 68.61 & 56.80 & 29.84 & 23.18 & 37.37 & 29.70 \\
        & InstanceRefer \cite{InstanceRefer} & PointGroup & ScanRefer & 3D & 77.82 & 66.69 & 34.57 & 26.88 & 44.27 & 35.80 \\
        & D3Net \cite{D3Net} & PointGroup & ScanRefer & 3D+2D & - & 70.35 & - & 30.05 & - & 37.87 \\
        & ViL3DRel \cite{ViL3DRel} & PointGroup & ScanRefer & - & 81.58 & 68.62 & 40.30 & 30.71 & 47.65 & 37.73 \\
        \cdashline{2-11}
        & ScanRefer \cite{ScanRefer} & VoteNet & ScanRefer & 3D+2D & 76.33 & 53.51 & 32.73 & 21.11 & 41.19 & 27.40 \\
        & SAT \cite{SAT} & VoteNet & ScanRefer & 3D+2D & 73.21 & 50.83 & 37.64 & 25.16 & 44.54 & 30.14 \\
        & FFL-3DOG \cite{FFL-3DOG} & VoteNet & ScanRefer & 3D & 78.80 & 67.94 & 35.19 & 25.70 & 41.33 & 34.01 \\
        & \multirow{2}{*}{3DVG-Trans \cite{3DVG-Transformer}} & \multirow{2}{*}{VoteNet} & \multirow{2}{*}{ScanRefer} & 3D & 77.16 & 58.47 & 38.38 & 28.70 & 45.90 & 34.47 \\
        & & & & 2D+3D & 81.93 & 60.64 & 39.30 & 28.42 & 47.57 & 34.67 \\
        & MVT \cite{MVTransformer} & VoteNet & ScanRefer & 3D & 77.67 & 66.45 & 31.92 & 25.26 & 40.80 & 33.26 \\
        & ViewRefer \cite{ViewRefer} & VoteNet & ScanRefer & 3D & 76.35 & 64.27 & 33.08 & 26.50 & 41.35 & 33.69 \\
        & 3D-SPS \cite{3D-SPS} & VoteNet & ScanRefer & 3D+2D & 84.12 & 66.72 & 40.32 & 29.82 & 48.82 & 36.98 \\
        & \multirow{2}{*}{3DJCG \cite{3DJCG}} & \multirow{2}{*}{VoteNet} & \multirow{2}{*}{ScanRefer} & 3D & 78.75 & 61.30 & 40.13 & 30.08 & 47.62 & 36.14 \\
        & & & & 3D+2D & 83.47 & 64.34 & 41.39 & 30.82 & 49.96 & 37.73 \\
        \Xhline{2.0\arrayrulewidth}
        \multirow{6}{*}{\rotatebox{90}{Pre-training}} & UniT3D \cite{UniT3D} & PointGroup & \tabincell{c}{Synthesize Data \\+ ScanRefer} & 3D & 82.75 & \underline{73.14} & 36.36 & 31.05 & 45.27 & 39.14 \\
        & 3D-VisTA \cite{3D-VisTA} & PointGroup & ScanScribe & 3D & 77.00 & 67.90 & 37.90 & 30.40 & 45.20 & 37.30 \\
        \cdashline{2-11}
        & \multirow{2}{*}{3D-VLP \cite{3D-VLP}} & \multirow{2}{*}{VoteNet} & \multirow{2}{*}{ScanRefer} & 3D & 79.35 & 62.60 & 42.54 & 32.18 & 49.68 & 38.08 \\
        & & & & 2D+3D & 84.23 & 64.61 & 43.51 & 33.41 & 51.41 & 39.46 \\
        \cdashline{2-11}
        & \multirow{2}{*}{\textbf{Ours}} & \multirow{2}{*}{VoteNet} & \multirow{2}{*}{ScanRefer} & 3D & 82.93 & 65.47 & 42.74 & 32.20 & 50.54 & 38.66 \\
        & & & & 2D+3D & \underline{84.67} & 66.38 & \underline{43.72} & \underline{33.52} & \underline{51.87} & \underline{39.91} \\
        \Xhline{2.0\arrayrulewidth}
    \end{tabular}}
    \caption{Comparison with state-of-the-art 3D VG methods combined with VoteNet or PointGroup on the ScanRefer validation set. ``Dataset'' denotes the (pre)-training dataset used in each model. The best results are underlined.}
    \label{table1}
\end{table*}

\subsection{Experimental settings}

\textbf{Datasets.} ScanRefer \cite{ScanRefer} is the earliest and most widely used dataset for 3D visual grounding (VG). It contains 51,583 free-form descriptions for 11,046 object instances from ScanNet \cite{ScanNet} indoor scenes. ScanRefer divides the data into the \emph{Unique} and \emph{Multiple} subsets according to the number of similar objects within the same category. We use the Scan2Cap \cite{Scan2Cap} dataset, built upon ScanRefer, to evaluate our captioning model. ScanQA \cite{ScanQA} is a newly released vision-language dataset for 3D question answering (QA), containing 43,363 questions and 32,337 unique answers. These question-answer pairs are generated from ScanRefer descriptions using a pre-trained question generation model.

\textbf{Evaluation metrics.} For 3D VG, we use $Acc@kIoU$ as the main metric, with $k$ typically set to 0.25 and 0.5. $Acc@kIoU$ represents the percentage of predicted boxes whose intersection-over-union (IoU) with the target object exceeds $k$. For 3D dense captioning, we adopt $m@kIoU$ to evaluate both localization accuracy and caption generation quality:
    $m@kIoU=\frac{1}{N}\sum_{i=1}^Nm(\hat{c}_i,c_i)\mathbb{1}(IoU(\hat{b}_i,b_i)\geq{k})$
, where $\hat{c_i}$ and $c_i$ represent the ground-truth and generated captions, respectively. $\hat{b}_i$ and $b_i$ denote the ground-truth and predicted boxes. $m$ includes caption metrics, such as BLEU-4 (B-4) \cite{BLEU}, CiDEr (C) \cite{CiDEr}, METEOR (M) \cite{METEOR} and ROUGE (R) \cite{ROUGE}. For 3D QA, we use the commonly used metric $EM@K$, with $K$ typically set to 1 and 10. $EM@K$ denotes the percentage of predictions where the top $K$ answers exactly match the ground-truth answer.

\textbf{Implementation details.} We adopt VoteNet \cite{VoteNet} as our detection module, setting the number of input point clouds $N$ and output 3D object proposals $M$ to 50,000 and 256, respectively. The feature dimensions $C_p$, $C_l$ are set to 256, and the number of graph convolutional layer is set to 3. For masked modality modeling, the mask ratios for input words and 3D object proposals are set to 0.2 and 0.75, respectively.

Our code is implemented using the Pytorch \cite{Pytorch} framework, and all experiments are conducted on a single NVIDIA A100 40G GPU \footnote{The model is pre-trained using nearly 60 hours.}. We follow 3D-VLP \cite{3D-VLP} to use the AdamW optimizer. The model is first pre-trained with a batch size of 16 for 200 epochs, with initial learning rates for the language encoder, detection module, and scene graph network set to 5e-4, 2e-3, and 5e-4, respectively. The model is then fine-tuned for 100, 100 and 30 epochs to adapt to 3D VG, 3D DC and 3D QA, respectively. For 3D VG and DC, the initial learning rate of task-specific head (\textit{e.g.}, grounding head or captioning head) is set to 5e-4, while the rest are set to 1e-4. For 3D QA, the initial learning rate is set to 1e-4 and decreased by 0.2 after 15 epochs. Details of downstream task fine-tuning can be found in the supplementary material.

\subsection{Downstream task results}

\textbf{3D visual grounding.} In Table \ref{table1}, we report the 3D VG results of our model combined with VoteNet detector. Our method, using both 2D and 3D inputs, achieves the best performance with 51.87\% at \textit{Overall} Acc@0.25 and 39.91\% at \textit{Overall} Acc@0.5, surpassing previous methods that employs the same VoteNet \cite{VoteNet} detector. Compared with the pre-training model 3D-VLP \cite{3D-VLP}, our approach shows a significant improvement of 0.46\% and 0.45\% at \textit{Overall} Acc@0.25 and Acc@0.5, respectively. Consistent improvements are also observed in both the \emph{Unique} and \emph{Multiple} subsets. This suggests that our method can effectively handle various types of scenes, benefiting from the universal features learned through our scene graph-guided contrastive learning scheme. Table \ref{table2} lists the results of several methods combined with more powerful detectors. Note that 3D-VisTA uses an offline 3D detector, while our model jointly optimizes the object detection module. We observe: (1) our model still outperforms previous methods with the same 3DETR \cite{3DETR} detector, but is slightly inferior to 3D-VisTA with Mask3D; (2) detection performance is positively correlated with VG performance. That is, using a powerful detector can lead to a notable improvement in VG performance.
\begin{table}[h]
    \centering
    \resizebox{\columnwidth}{!}{
    \begin{tabular}{r|c|cc}
        \Xhline{2.0\arrayrulewidth}
        Method & Detector & Acc@0.25 & Acc@0.5 \\
        \Xhline{2.0\arrayrulewidth}
        BUTD-DETR \cite{BUTD-DETR} & 3DETR-like \cite{3DETR} & 49.76 & 37.05 \\
        EDA \cite{EDA} & 3DETR \cite{3DETR} & \underline{53.83} & 41.70 \\
        3D-VisTA \cite{3D-VisTA} & Mask3D \cite{Mask3D} & 50.60 & \underline{45.80} \\
        \cdashline{1-4}
        \textbf{Ours} & 3DETR \cite{3DETR} & 53.69 & 42.04 \\
        \Xhline{2.0\arrayrulewidth}
    \end{tabular}}
    \caption{Comparison with several 3D VG methods combined with more powerful detector. 3D-VisTA uses an offline 3D detector, while other methods jointly optimize the object detection module.}
    \label{table2}
\end{table}

\textbf{3D question answering.} Table \ref{table3} lists the quantitative results of different approaches on 3D QA. ``MCAN'' refers to the modular co-attention network \cite{MCAN}. Our method achieves the best results, with 24.80\% at EM@1 and 59.24\% at EM@10, outperforming 3D-VLP \cite{3D-VLP} by 3.15\% and 8.78\% in terms of EM@1 and EM@10, respectively. In addition, our method shows a significant improvement (i.e., 2.54\% at EM@1 and 4.73\% at EM@10) over the task-specific FE-3DGQA \cite{FE-3DGQA}. This further demonstrates that the 3D-language features learned through our pre-training scheme are semantically aligned and enhanced in granularity.
\begin{table}[h]
    \centering
    \resizebox{0.8\columnwidth}{!}{
    \begin{tabular}{r|cc}
        \Xhline{2.0\arrayrulewidth}
        Method & EM@1 & EM@10 \\
        \Xhline{2.0\arrayrulewidth}
         VoteNet \cite{VoteNet} + MCAN \cite{MCAN} & 17.33 & 45.54 \\
         ScanRefer \cite{ScanRefer} + MCAN \cite{MCAN} & 18.59 & 46.76 \\
         ScanQA \cite{ScanQA} & 20.28 & 50.01 \\
         FE-3DGQA \cite{FE-3DGQA} & 22.26 & 54.51 \\
         3D-VLP \cite{3D-VLP} & 21.65 & 50.46 \\
         3DVLP \cite{3DVLP} & 24.03 & 57.91 \\
         \cdashline{1-3}
         \textbf{Ours} & \underline{24.80} & \underline{59.24} \\
         \Xhline{2.0\arrayrulewidth}
    \end{tabular}}
    \caption{Comparison with state-of-the-art 3D question answering methods on the ScanQA validation set. The best results are underlined.}
    \label{table3}
\end{table}

\textbf{3D dense captioning.} In Table \ref{table4}, we present the quantitative results on 3D DC. We observe that our method, using both 2D and 3D inputs, achieves best results across captioning metrics, significantly outperforming task-specific methods and performing comparably to other pre-training approaches. Specifically, compared with 3D-VLP \cite{3D-VLP}, our method shows slight improvements of 0.43\% at B-4@0.5 and 0.72\% at M@0.5. Additionally, compared to the task-specific 3DJCG \cite{3DJCG}, our approach achieves notable gains of 1.71\% and 1.33\% at B-4@0.5 and M@0.5, respectively. This indicates that our generated captions are more descriptive and closer to human expression. This is mainly due to our proposed pre-training scheme, which helps learn highly transferable 3D-language features, enabling our fine-tuned caption model to generate more accurate descriptions based on 3D object features.
\begin{table}[h]
    \centering
    \resizebox{\columnwidth}{!}{
    \begin{tabular}{c|r|c|cccc}
        \Xhline{2.0\arrayrulewidth}
        & Method & Input & C@0.5 & B-4@0.5 & M@0.5 & R@0.5 \\
        \Xhline{2.0\arrayrulewidth}
        \multirow{12}{*}{\rotatebox{90}{Task-specific}} & \multirow{2}{*}{Scan2Cap \cite{Scan2Cap}} & 3D & 35.20 & 22.36 & 21.44 & 43.57 \\
        & & 2D+3D & 39.08 & 23.32 & 21.97 & 44.48 \\
        & \multirow{2}{*}{X-Trans2Cap \cite{X-Trans2Cap}} & 3D & 41.52 & 23.83 & 21.90 & 44.97 \\
        & & 2D+3D & 43.87 & 25.05 & 22.46 & 45.28 \\
        & \multirow{2}{*}{MORE \cite{MORE}} & 3D & 38.98 & 23.01 & 21.65 & 44.33 \\
        & & 2D+3D & 40.94 & 22.93 & 21.66 & 44.42 \\
        & D3Net \cite{D3Net} & 2D + 3D & 47.32 & 24.76 & 21.66 & 43.62 \\
        & SpaCap3D \cite{SpaCap3D} & 2D+3D & 44.02 & 25.26 & 22.33 & 45.36 \\
        & REMAN \cite{REMAN} & 2D+3D & 45.00 & 26.31 & 22.67 & 46.96 \\
        & Contextual \cite{Contextual} & 2D+3D & 46.11 & 25.47 & 22.64 & 45.96 \\
        & \multirow{2}{*}{3DJCG \cite{3DJCG}} & 3D & 50.02 & 31.87 & 24.53 & 51.17 \\
        & & 2D+3D & 49.48 & 31.03 & 24.22 & 50.80 \\
        \Xhline{2.0\arrayrulewidth}
        \multirow{5}{*}{\rotatebox{90}{Pre-training}} & UniT3D \cite{UniT3D} & 3D & 46.69 & 27.22 & 21.91 & 45.98 \\
        & \multirow{2}{*}{3D-VLP \cite{3D-VLP}} & 3D & 50.02 & 31.87 & 24.53 & 51.17 \\
        & & 2D+3D & 54.94 & 32.31 & 24.83 & 51.51 \\
        \cdashline{2-7}
        & \multirow{2}{*}{\textbf{Ours}} & 3D & 52.60 & 32.49 & 24.93 & 51.44 \\
        & & 2D+3D & \underline{55.32} & \underline{32.74} & \underline{25.55} & \underline{52.58} \\
        \Xhline{2.0\arrayrulewidth}
    \end{tabular}}
    \caption{Comparison with competitive 3D dense captioning methods on the Scan2Cap dataset. The best results are underlined.}
    \label{table4}
\end{table}

\begin{figure*}[t]
    \centering
    \includegraphics[width=\linewidth]{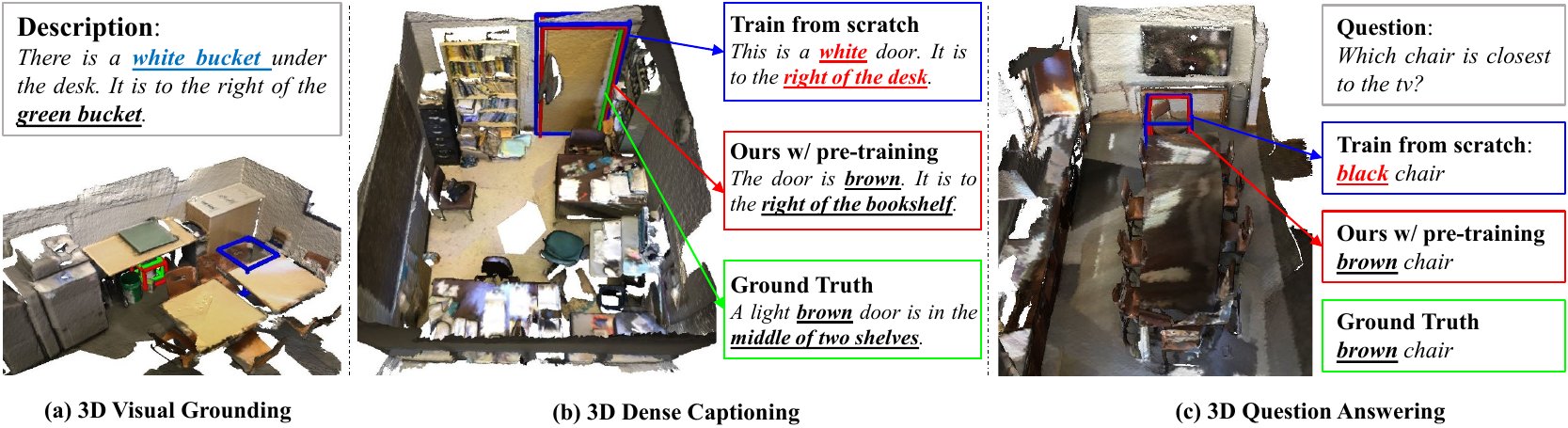}
    \caption{Qualitative results on downstream tasks: (a) 3D visual grounding, (b) 3D dense captioning and (c) 3D question answering. The \textcolor{green}{green box} indicates the \textcolor{green}{ground truth}, the \textcolor{blue}{blue box} represents predictions from our model \textcolor{blue}{trained from scratch}, and the \textcolor{red}{red box} shows predictions from our \textcolor{red}{pre-training model}.} 
    \label{fig_5}
\end{figure*}

\textbf{Qualitative results.} Figure \ref{fig_5} presents our qualitative results on 3D VG, 3D DC and 3D QA. We observe that our pre-training model achieves more accurate bounding boxes, generates more descriptive captions, and produces answers that align better with human consensus compared to training from scratch. More qualitative results are provided in the supplementary material.

\begin{table*}[t]
    \centering
    \resizebox{0.8\textwidth}{!}{
    \begin{tabular}{l|cc|cccc|cc}
        \Xhline{2.0\arrayrulewidth}
        \multirow{2}{*}{Method} & \multicolumn{2}{c|}{3D VG} & \multicolumn{4}{c|}{3D DC} & \multicolumn{2}{c}{3D QA} \\
        & Acc@0.25 & Acc@0.5 & C@0.5 & B-4@0.5 & M@0.5 & R@0.5 & EM@1 & EM@10 \\
        \Xhline{2.0\arrayrulewidth}
        Scratch & 49.85 & 37.49 & 52.14 & 30.98 & 24.31 & 50.42 & 23.31 & 58.56 \\
        + SG\_MCL & 50.82 & 38.26 & 53.56 & 31.50 & 24.43 & 51.44 & 24.45 & 58.93 \\
        + MMM & 51.06 & 38.85 & 54.50 & 31.97 & 24.83 & 51.75 & 24.37 & 58.86 \\
        \textbf{+ SG\_MCL + MMM} & \underline{51.87} & \underline{39.91} & \underline{55.32} & \underline{32.74} & \underline{25.55} & \underline{52.58} & \underline{24.80} & \underline{59.24} \\
        \Xhline{2.0\arrayrulewidth}
    \end{tabular}}
    \caption{Ablation study on the pre-training objectives.}
    \label{table5}
\end{table*}

\subsection{Ablation study}
In this section, we conduct extensive ablation experiments to validate the design choices of our 3D vision-language framework. We use the fine-tuned model on 3D visual grounding task for evaluation due to its simplicity.

\textbf{Ablation on pre-training objectives.} Here, we analyze the contributions of different pre-training objectives. For a more comprehensive evaluation, we gradually add our pre-training objectives. Table \ref{table5} presents the results of our model on 3D VG, 3D DC and 3D QA under various combinations of pre-training objectives. Notice that, ``Scratch'' refers to directly training the model on down-steam tasks without pre-training. We observe that our model, with the proposed pre-training objectives, significantly improves downstream performance. Specifically, scene graph-guided multi-level contrastive learning (SG\_MCL) achieves substantial gains of 0.97\% and 0.77\% at VG Acc@0.25 and Acc@0.5, respectively. Consistent gains are also observed on captioning (1.42\% at C@0.5 and 0.52\% at B-4@0.5) and QA (1.14\% at EM@1 and 0.37\% at EM@10) metrics. This demonstrates the effectiveness of our scene graph-based multi-level contrastive learning and masked modality modeling.

\textbf{Ablation on the number of scene graph layers.} We investigate the impact of the number of scene graph layers on downstream VG performance. As shown in Table \ref{table6}, as the depth of scene graph network increases, the VG performance slightly improves. This is because that more layers enable nodes (objects) and edges in the scene graph to better capture contextual information from neighboring nodes, which enhances sentence-referred object alignment and universal feature learning. However, considering the computational cost, we set the number of scene graph layers to 3.
\begin{table}[h]
    \centering
    \resizebox{0.75\columnwidth}{!}{
    \begin{tabular}{c|cccc}
        \Xhline{2.0\arrayrulewidth}
        \# SG layer & 1 & 2 & 3 & 4 \\
        \Xhline{2.0\arrayrulewidth}
        Acc@0.25 & 50.74 & 51.36 & 51.87 & \underline{52.01} \\
        Acc@0.5 & 38.56 & 39.48 & \underline{39.91} & 39.85 \\ 
        \Xhline{2.0\arrayrulewidth}
    \end{tabular}}
    \caption{Ablation study on the number of scene graph layers.}
    \label{table6}
\end{table}

\textbf{Ablation on the type of scene graph layer.} We further examine the impact of different types of scene graph layers on downstream VG performance, as shown in Table \ref{table7}. ``GCN'' refers to the graph convolutional layer, while ``EdgeConv'' denotes the edge convolutional layer \cite{EdgeGCN}. We found that the strong representation capability of the EdgeConv layer enhances VG performance improvement, leading us to select EdgeConv as our base scene graph layer. This improvement may be attributed to the fact that nodes with enhanced context-learning capabilities can achieve more precise 3D object-text alignment. 
\begin{table}[h]
    \centering
    \resizebox{0.65\columnwidth}{!}{
        \begin{tabular}{c|cc}
        \Xhline{2.0\arrayrulewidth}
        Method & Acc@0.25 & Acc@0.5 \\
        \Xhline{2.0\arrayrulewidth}
        GCN & 51.25 & 39.45 \\
        EdgeConv \cite{EdgeGCN} & \underline{51.87} & \underline{39.91} \\
        \Xhline{2.0\arrayrulewidth}
    \end{tabular}}
    \caption{Ablation study on the type of scene graph layer.}
    \label{table7}
\end{table}
\section{Conclusion}

In this paper, we leverage the connection between 3D scene graphs and natural language, proposing a 3D scene graph-guided vision-language pre-training framework. Our method uses simple modality encoders, graph convolutional layers, and cross-attention layers to learn universal representations with strong transferability, avoiding the need for complex, task-specific designs. We align 3D objects and textual features via the proposed scene graph-guided contrastive learning and masked modality learning. Through extensive experiments, our pre-training model can be fine-tuned to adapt to various downstream tasks, achieving performance comparable to or better than existing methods. However, it remains challenging to pre-train our model on 3D-text pairs collected from different types of sensors due to significant differences in point cloud density. In the future, we will explore cross-domain generalization for 3D vision-language pre-training.

{
    \small
    \bibliographystyle{ieeenat_fullname}

}

\clearpage
\appendix
\renewcommand{\thesection}{A.\arabic{section}}
\renewcommand{\thesubsection}{A.\arabic{section}.\arabic{subsection}}

\section{Downstream task fine-tuning}

\textbf{3D visual grounding (VG)} aims to identify the target object that best matches a given language description. Following common practice, we treat 3D VG as a classification problem, adding three MLP layers on top of the multi-modal features to predict grounding scores. The target label is a multi-hot label. We assign 1 to those object proposals whose IoU with the referential target is greater than 0.25. The model is fine-tuned using softmax cross-entropy loss.

\textbf{3D dense captioning (DC)} requires the model to detect and describe objects within a 3D scene. To achieve this, we add a captioning module similar to Scan2Cap \cite{Scan2Cap} on top of object (node) features, which takes the 1$^{st}$ to the ($L$-1)$^{th}$ words as input and predicts the next word in an auto-regressive manner. The model is fune-tuned using cross entropy loss.

\textbf{3D question answering (QA)} requires the model to answer a given question and locate question-relevant objects within a 3D scene. We use a modular co-attention network (MCAN) \cite{MCAN} to fuse object proposal and text features to predict the answer. The model is fine-tuned using binary cross entropy loss and a question-object contrastive loss.

\section{Model architecture}
For the scene encoder, we use a PointNet++ \cite{PointNet++} network with four set abstraction (SA) layers and two feature propagation (FP) layers to extract 256-dimensional features with 1024 points. The radii of the four SA layers are set to 0.2, 0.4, 0.8 and 1.2, respectively. The voting module proposed by \cite{VoteNet} is then used to aggregate seed points and generate 256 object proposals. For the text encoder, we use a GRU \cite{GRU} module with pre-trained GloVE \cite{GloVE} embeddings to extract 256-dimensional textual features. The feature dimension is set to 256 in the subsequent cross-modality fusion layers. For the scene graph network, we employ a three-layer EdgeConv \cite{EdgeGCN} to update node features, enabling each node to incorporate contextual information from its neighbors. After obtaining the object and textual features, we use two cross-attention layers to extract multi-modal features.

\subsection{More qualitative results}
\begin{figure*}[h]
    \centering
    \includegraphics[width=0.85\textwidth]{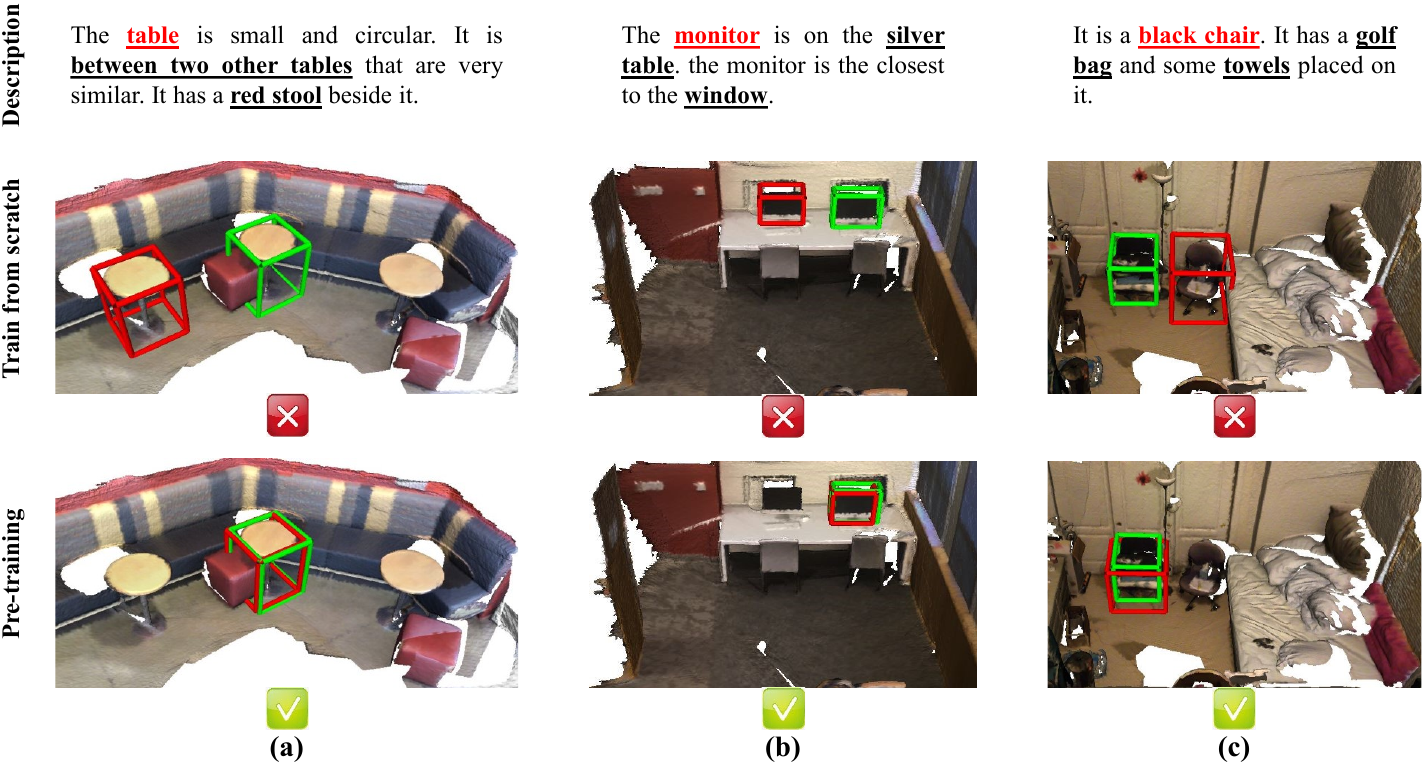}
    \caption{Qualitative results on downstream 3D visual grounding. The \textcolor{green}{Green box} represents the ground-truth, and the \textcolor{red}{red box} indicates the prediction.\label{Supp_VG}}
\end{figure*}

\begin{figure*}[h]
    \centering
    \includegraphics[width=\textwidth]{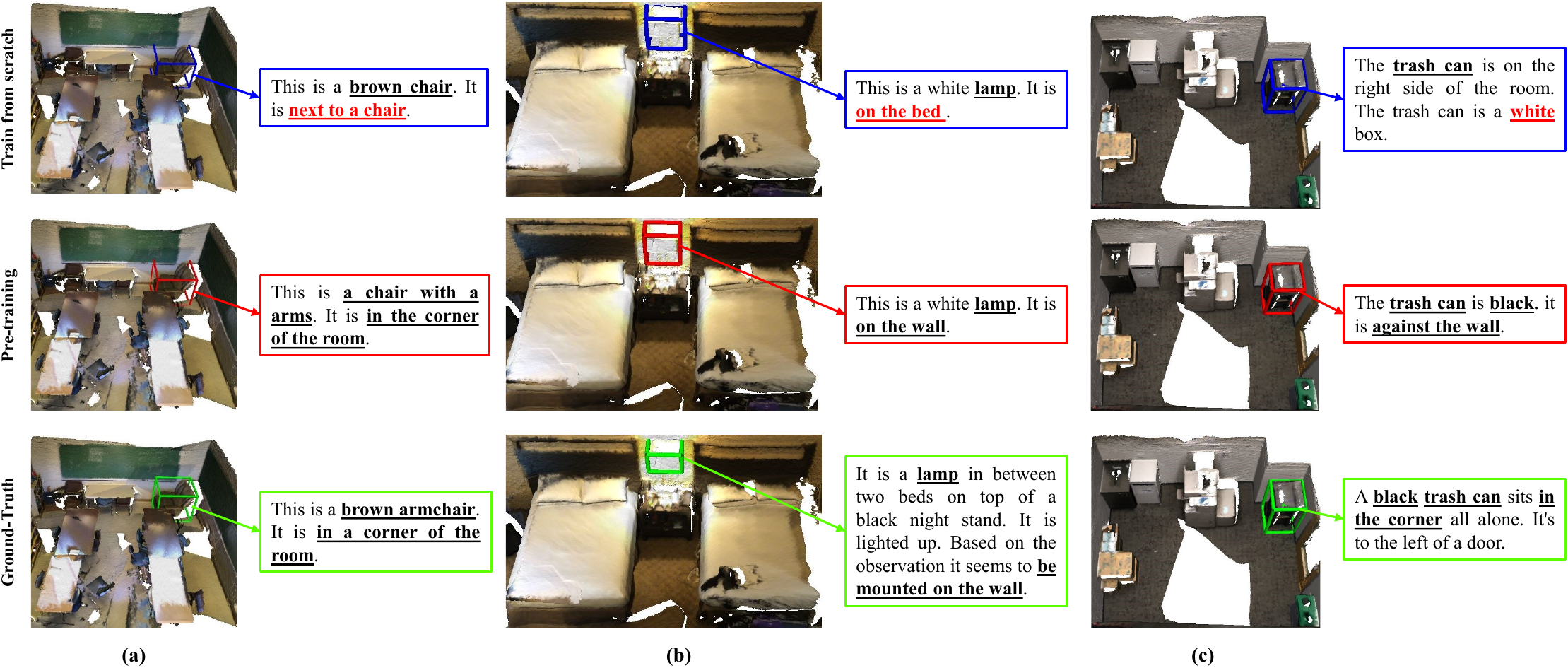}
    \caption{Qualitative results on downstream 3D dense captioning. \textcolor{green}{Green box} for the ground-truth, and \textcolor{red}{red box} for the prediction. The accurate parts of generated captions that match ground-truth are underlined and the inaccurate parts are in red.\label{Supp_DC}}
\end{figure*}

\begin{figure*}[h]
    \centering
    \includegraphics[width=0.85\textwidth]{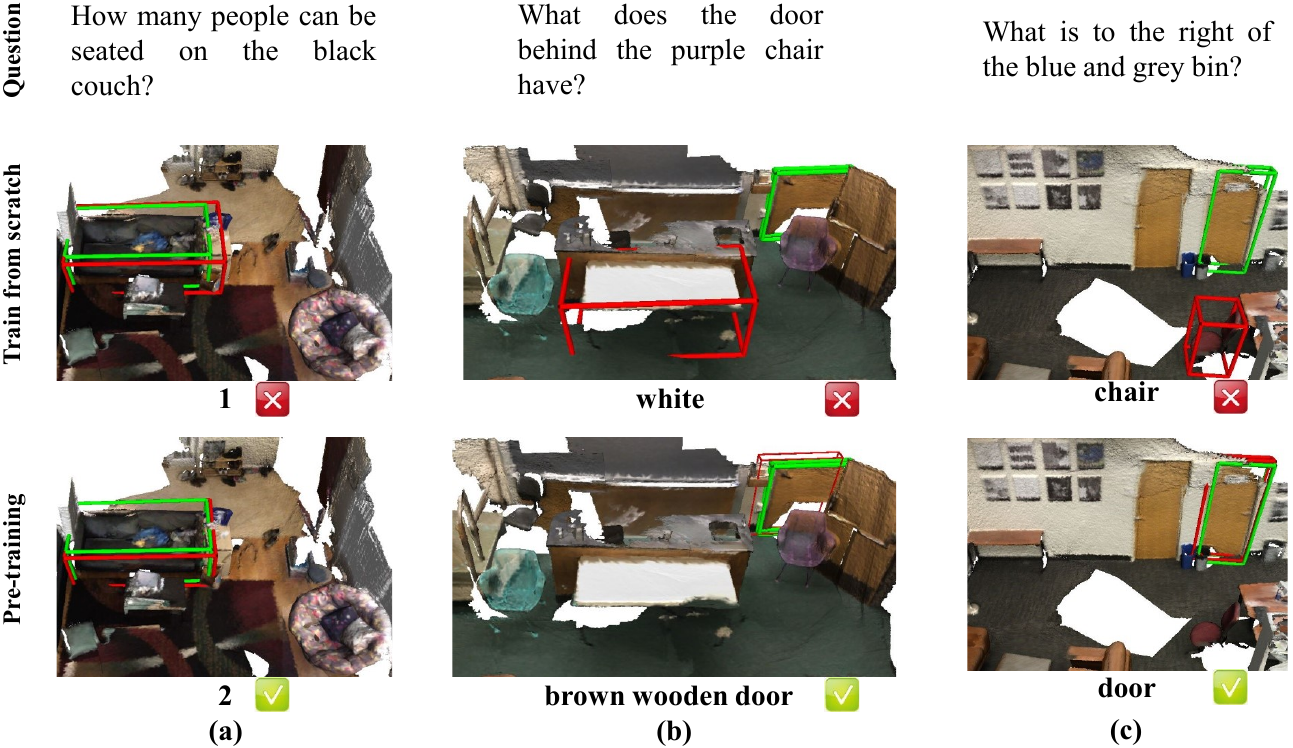}
    \caption{Qualitative results on downstream 3D question answering. The \textcolor{green}{green box} represents the ground-truth, and the \textcolor{red}{red box} indicates the predictions. The top 1 predicted answers are reported. \label{Supp_QA}}
\end{figure*}

We provide additional qualitative results for 3D visual grounding (Fig. \ref{Supp_VG}), 3D dense captioning (Fig. \ref{Supp_DC}) and 3D question answering (Fig. \ref{Supp_QA}). In Fig. \ref{Supp_VG}, our method achieves more accurate localization results than training from scratch on the ScanRefer dataset, particularly when similar objects are present in the scene. This shows that our method is capable of fully understanding the content of the given text. In Fig. \ref{Supp_DC}, our method effectively recognizes spatial relationships between objects and generates accurate and descriptive captions. Figure \ref{Supp_QA} demonstrates that our pre-training scheme enhances the model’s performance in both localizing target bounding boxes and answering questions.

\begin{figure*}[h]
    \centering
    \includegraphics[width=0.6\textwidth]{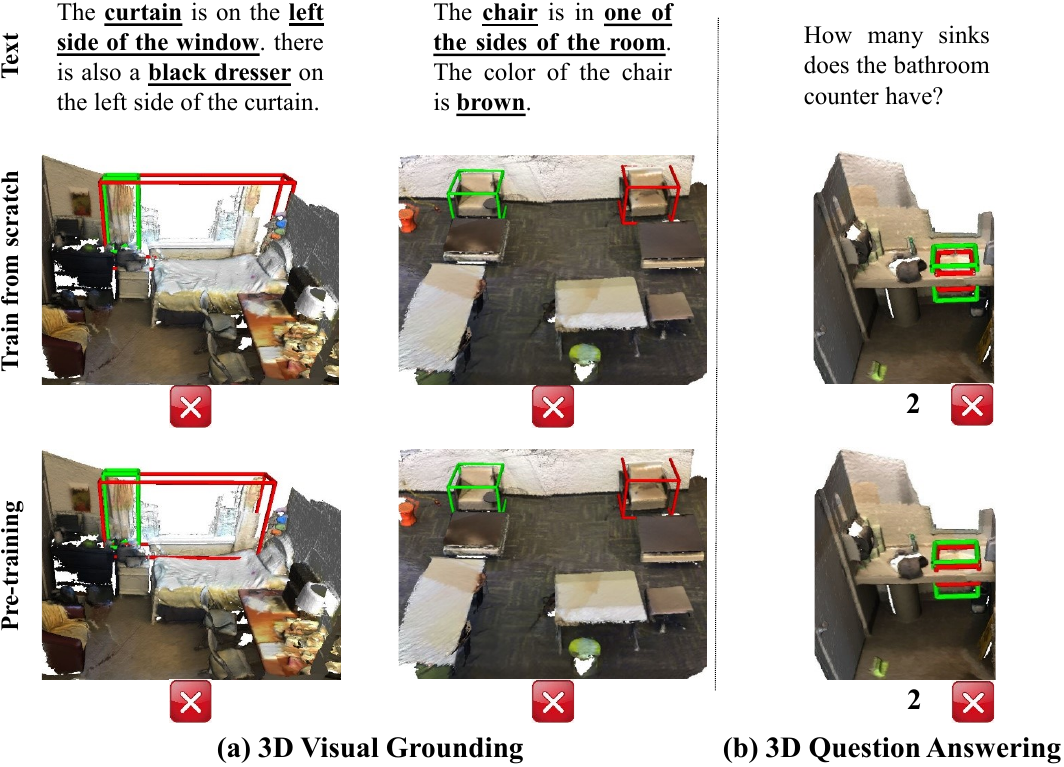}
    \caption{Failure cases on downstream tasks: (a) 3D visual grounding and (b) 3D question answering.\label{failures}}
\end{figure*}

\subsection{Failure cases}
Figure \ref{failures} presents several failure examples on 3D visual grounding and 3D question answering. We observe the following issues: (1) For the visual grounding example in the first column, our model still struggles to understand complex cases involving spatial relations. In the future, we can further introduce edge (object relation) constraints to handle this problem. (2) For the visual grounding example in the second column, the input text is inherently ambiguous, meaning multiple objects in the scene could match the description, leading to failure due to the ambiguity in the text. This may be due to the dataset itself, as it contains multiple ambiguous descriptions. That is, one ground truth description corresponds to multiple similar objects in the scene. (3) For the question answering example in the third column, our method has difficulty answering counting questions, such as counting the number of related targets in the scene. This is actually a common issue with existing vision-language models, as they primarily focus on modeling the geometry and appearance of objects.


\end{document}